\documentclass[10pt,twocolumn,letterpaper]{article}

\usepackage{iccv}
\usepackage{times}
\usepackage{epsfig}
\usepackage{graphicx}
\usepackage{amsmath}
\usepackage{amssymb}

\usepackage{color}
\usepackage{booktabs}
\usepackage{enumitem}
\usepackage{array}
\usepackage{float}
\usepackage{subcaption}
\usepackage{appendix}
\usepackage{multirow}

\usepackage[pagebackref=true,breaklinks=true,bookmarks=false, letterpaper=true,colorlinks]{hyperref}

\iccvfinalcopy 

\def\iccvPaperID{8118} 

\ificcvfinal\fi

\begin{document}

\title{Minimal Adversarial Examples for Deep Learning on 3D Point Clouds}

\author{Jaeyeon Kim$^1$ \quad\quad Binh-Son Hua$^{2,3}$ \quad\quad Duc Thanh Nguyen$^4$ \quad\quad Sai-Kit Yeung$^{1}$
	\vspace{0.2cm}\\
	$^1$Hong Kong University of Science and Technology \quad\quad\quad $^2$VinAI Research, Vietnam \\
	$^4$Deakin University \quad\quad\quad\quad\quad\quad\quad\quad\quad\quad\quad\quad\quad\quad\quad\quad
	$^3$VinUniversity, Vietnam
	\vspace{-0.2cm}\\
}

\maketitle
\ificcvfinal\thispagestyle{empty}\fi

\begin{abstract}
   With recent developments of convolutional neural networks, deep learning for 3D point clouds has shown significant progress in various 3D scene understanding tasks, e.g., object recognition, semantic segmentation. In a safety-critical environment, it is however not well understood how such deep learning models are vulnerable to adversarial examples. In this work, we explore adversarial attacks for point cloud-based neural networks. We propose a unified formulation for adversarial point cloud generation that can generalise two different attack strategies. Our method generates adversarial examples by attacking the classification ability of point cloud-based networks while considering the perceptibility of the examples and ensuring the minimal level of point manipulations. Experimental results show that our method achieves the state-of-the-art performance with higher than 89\% and 90\% of attack success rate on synthetic and real-world data respectively, while manipulating only about 4\% of the total points.
\end{abstract}

\section{Introduction}

\noindent Deep learning has shown great potentials in solving a wide spectrum of computer vision tasks. In life-crucial applications, one concern is that deep neural networks can be vulnerable to adversarial examples, a special kind of inputs that can fool the networks to make undesirable predictions. Several adversarial attack techniques have been proposed to generate such examples. In contrast, adversarial defense methods have been developed to detect and neutralise adversarial examples. Therefore, understanding how adversarial attacks and defenses operate is of great importance to make deep learning techniques more reliable and robust. 

With the growing popularity of low-cost 3D sensors and light-field cameras, the community has started investigating the vulnerability of deep networks on 3D data, especially 3D point clouds~\cite{chong-first-adv-2019,Daniel-attack-and-defense1-2019,Daniel-attack-and-defense2-2019,Jiancheng-bidu-2019,Yuxin-Geometry-2019}. However, existing works focus on common scenarios, such as generating adversarial point clouds by perturbing points. While such approaches have high attack success rates, the perturbations are not imperceptible and can be identified easily by outlier detection or noise removal algorithms. In addition, existing adversarial attack methods do not perform optimally since all points in a point cloud are involved in the manipulation.

In this work, we study 3D adversarial attack in a more extreme but practical setting: how to generate an adversarial point cloud with minimum number of points perturbed from an original point cloud while maintaining the perceptibility of the original point cloud (see Figure~\ref{fig:best_case}). To address this problem, we propose a new formulation for adversarial point cloud generation that can be adapted to different attack strategies. The novelty of our work lies in both the research problem and proposed solution. Specifically, minimal 3D point cloud attack is an unexplored problem. We are also the first to propose a formulation that (i) considers both the perceptibilty and optimality of adversarial samples, and (ii) generalises both point perturbation and addition in a unified framework. In summary, our contributions include:

\begin{figure}[t!]
\centering
\includegraphics[width=1.0\linewidth]{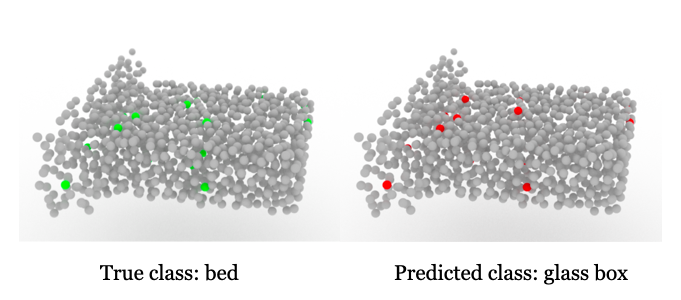}
\caption{Left: input point cloud (1,024 points) classified correctly by PointNet~\cite{qi2017pointnet} and 15 selected points (in green). Right: adversarial point cloud misclassified by PointNet and perturbed locations (in red).}
\label{fig:best_case}
\end{figure}

\begin{itemize}[leftmargin=*]

\item A new technique to generate minimal 3D adversarial point clouds;

\item A unified formulation that generalises two adversarial point cloud generation strategies: point perturbation and point addition;

\item A vulnerability analysis on the relation between the perturbed points found by our method and the concept of critical points in PointNet~\cite{qi2017pointnet};

\item A benchmark of adversarial attacks on both synthetic and real-world 3D point clouds, which shows our method achieves consistent performance over both domains.
 
\end{itemize}

\section{Related Work}

\paragraph{3D Deep Learning.} 
Recent availability of 3D datasets~\cite{Zhirong-shapenets-2014,Angel-shapenet-2015,hua2016scenenn,armeni-parsing-cvpr16,dai2017scannet,chang-matterport3d-3dv17,Mikaela-real-world-dataset-2019} has led to significant advances in deep learning on 3D data. In this domain, most existing works focus on designing convolution operations that enable convolutional neural networks to learn features directly from point clouds~\cite{qi2017pointnet,qi2017pointnet++,li2018pointcnn,xu-spidercnn-eccv2018,hua2017point,wang2018edgeconv,shen2018mining,huang2018recurrent,zhang-shellnet-iccv19}. Several attempts have been made to create rotation invariant convolutions~\cite{rao-spherical-cvpr19,zhang-riconv-3dv19,poulenard-spherical-3dv19,you-prin-aaai20}. Such convolutions allow scene understanding tasks, e.g., object recognition, semantic segmentation, to be trained directly with various input point clouds. Other improvements have also been found in the literature. For instance, Liu et al.~\cite{liu2019point2sequence} proposed an  RNN-based model to extract the correlations of local areas. Yan et al.~\cite{yan2020pointasnl} extended the pointwise MLP network by adopting adaptive sampling to handle outilers and noise.


In this work, we adopt PointNet~\cite{qi2017pointnet} as a target neural network to investigate important aspects of our method due to its popularity. However, we also verify our method with different deep learning models including  PointNet++~\cite{qi2017pointnet++}, DGCNN~\cite{wang2018edgeconv}, SpiderCNN~\cite{xu-spidercnn-eccv2018}, and PointASNL~\cite{yan2020pointasnl}.

\paragraph{Adversarial Point Clouds.}
There exist studies on adversarial attacks and defenses for point cloud classification~\cite{chong-first-adv-2019,Daniel-attack-and-defense1-2019,Daniel-attack-and-defense2-2019,Jiancheng-bidu-2019,Zhao_isometry_2020cvpr, tsai2020robust,Hamdi-AdvPC-eccv2020, Zhou_LGGAN_2020cvpr, lee2020shapeadv}. For instance, Xiang et al.~\cite{chong-first-adv-2019} proposed an algorithm for point perturbation and addition based on the attack framework in~\cite{Nicholas-Towards-2017} using Chamfer and Hausdorff distance. Liu et al.~\cite{Daniel-attack-and-defense1-2019} extended the fast gradient sign method in~\cite{goodfellow-fgsm-iclr14} for constructing 3D adversarial examples using mesh and clipping norm. The basic ideas in these works follow previous adversarial attack techniques in 2D domain, which focus on how a point cloud should be perturbed to make an adversarial example. Readers are referred to~\cite{yuan-adversarial-example-nn19} for a comprehensive review on adversarial attacks and defenses on images, and to~\cite{Ilyas_not_bugs_NEURIPS2019} for a general analysis and proof of existence of adversarial examples.

It is also possible to mix newly added points with perturbed points in a point cloud to make adversarial examples~\cite{Daniel-attack-and-defense2-2019}. Yuxin et al.~\cite{Yuxin-Geometry-2019} considered the consistency of local curvatures in a loss function to guide perturbed points lean towards object surfaces. Tsai et al.~\cite{tsai2020robust} incorporated the K-Nearest Neighbor loss in ~\cite{chong-first-adv-2019} to constrain adversarial samples to become physical objects. Zhao et al.~\cite{Zhao_isometry_2020cvpr} showed the vulnerability of isometry transformation by making perturbations on isometry matrix. Hadmi et al.~\cite{Hamdi-AdvPC-eccv2020} applied auto-encoders in perturbation to improve the transferability of adversarial examples across networks. Lee et al.~\cite{lee2020shapeadv} added perturbation noise into the latent space of auto-encoders to maintain the shape of input point clouds. Zhou et al.~\cite{Zhou_LGGAN_2020cvpr} employed GAN~\cite{Goodfellow_GAN_NIPS2014} in generating adversarial point clouds with predefined target labels. 

In addition to creating adversarial examples by perturbing existing points or adding new points into point clouds, one can fool classification techniques by removing points from input data~\cite{Zheng-saliency-map-2019,Matthew-Robustness-map-2019,Jiancheng-bidu-2019}. For instance, Zheng et al.~\cite{Zheng-saliency-map-2019} eliminated important points on saliency map of an input object. Matthew et al.~\cite{Matthew-Robustness-map-2019} iteratively removed critical points from an input point cloud until a target classification technique failed to classify the point cloud.

In contrast to adversarial attacks, countermeasures for adversarial point clouds have been so far scarce. Typical defense approaches include outlier or salient point removal~\cite{Daniel-attack-and-defense1-2019} and noise removal~\cite{Jiancheng-bidu-2019,Zhou-2019-ICCV}. Recently, Dong et al.~\cite{Dong-Gather-Vector-cvprR20} used relative position of each local part of a clean point cloud to global object center as adversarial indicator. Wu et al.~\cite{wu-ifdefense-arxiv2020} proposed a method to predict implicit functions capturing clean shapes of point clouds. 

\paragraph{Minimal Adversarial Attacks.}
In 2D domain, there is a specific family of techniques that focus on perturbing a minimum number of pixels in adversarial attacks. 

For instance, Papernot et al.~\cite{nicolas_JSMA_2016} perturbed pixels on saliency maps. Carlini et al.~\cite{Nicholas-Towards-2017} extended this method and used $\ell_0$-norm optimisation to minimise the number of pixels to perturb. Recently, Modas et al.~\cite{Modas_sparsefool_2019} and Croce et al.~\cite{Croce_sparse2_2019} focused on how to perturb a sparse set of pixels while still achieving good perceptibility. Local search and evolutionary algorithms were also applied to obtain sparse perturbations in~\cite{Nina_localsearch_2017,Jiawei_one_2019,Lukas_ev2_2019}. 

In this paper, we also explore adversarial attacks that only manipulate a minimal set of points. However, unlike the above works, we propose a new formulation that is general and can be adapted to various adversarial point cloud generation strategies. In addition, we also consider the perceptibility of adversarial examples in our formulation.
\section{Proposed Method}
\label{sec:ProposedMethod}

\noindent Our problem of interest can be stated as follows. Let $P=\left\{\mathbf{p}_1,...,\mathbf{p}_{N} \right\}$ be an input set of $N$ points where each point $\mathbf{p}_i$ is represented by a vector of its coordinates $\mathbf{p}_i=\left[p_{i,x}, p_{i,y}, p_{i,z}\right]^{\top} \in \mathbb{R}^3$. Let $F$ denote a point-based neural network, e.g., PointNet~\cite{qi2017pointnet}, and $F_i(P)$ denote the probability that the point set $P$ is classified into the $i$-th class. Ideally, if $i^*$ is the true class label of the point cloud $P$, then $i^*=\operatorname{argmax}_{i} F_i(P)$. Let $P'$ be an adversarial example generated from $P$. We aim to find $P'$ that satisfies the following conditions:
\begin{itemize}
\item[(i)] The perceptibility of $P$ is maintained, i.e., the generated point cloud $P'$ should not much deform from $P$;
\item[(ii)] A minimum number of points in $P$ are manipulated;
\item[(iii)] $\operatorname*{argmax}_i F_i(P) \neq \operatorname*{argmax}_{i'} F_{i'}(P')$, i.e., $P'$ and $P$ are classified into different classes by the network $F$. 
\end{itemize}

Note that generation of adversarial point clouds by compromising the attack success rate and perceptibility of adversarial samples has been investigated in the literature, e.g.,~\cite{chong-first-adv-2019}. However, our work differs from~\cite{chong-first-adv-2019} in the following points. First, our focus
is untargeted attack while that is targeted attack in~\cite{chong-first-adv-2019}.
Second, the number of manipulated points is not considered in~\cite{chong-first-adv-2019}, leading to extremely high numbers of points are manipulated (as shown in our experiments). Third, while critical points proposed in PointNet~\cite{qi2017pointnet} are used to drive the solution in~\cite{chong-first-adv-2019}, we show that our method could reach those critical points yet occupy a small portion, proving the capability of our method of finding compact yet vulnerable point sets.

In the following, we present a new formulation to generate $P'$ using $\ell_0$-norm optimisation and describe in details how our formulation can be applied to point perturbation and point addition.


\subsection{Point Perturbation}
\label{sec:PointPerturbation}

\subsubsection{Formulation} 

Given a point cloud $P$, we aim to find a minimal set of points that can be shifted to generate an adversarial point cloud $P'$ to attack the network $F$. We express the selection of points in $P$ for perturbation by a binary indication vector $\mathbf{a} = \left[a_1,...,a_N\right]^{\top} \in \{0,1\}^N$ where $a_i$ is 1 if $\mathbf{p}_i$ is selected, and 0 otherwise. Suppose that $E = \{\mathbf{e}_1,...,\mathbf{e}_N\}$ is the set of perturbations, in which $\mathbf{e}_i=\left[e_{i,x}, e_{i,y}, e_{i,z}\right]^{\top} \in \mathbb{R}^3$ is the perturbation vector to be applied on $\mathbf{p}_i$ to obtain $\mathbf{p}_i'$. Applying perturbation set $E$ on the point cloud $P$ results in an adversarial point cloud $P'$ as
\begin{equation} 
\label{eq_adversarialpointcloud}
P'=\left\{\mathbf{p'}_i = \mathbf{p}_i + a_i \mathbf{e}_i \mid \mathbf{p}_i \in P\right\} \,.
\end{equation}

The process of generation of $P'$ can be formulated as,
\begin{equation}
\begin{aligned}
\label{eq_pointperturbation_1}
&\min_{\mathbf{a}, E} f(P,\mathbf{a}, E) = \min_{\mathbf{a}, E} \left\{ \lambda_1 \| \mathbf{a} \|_0 + \lambda_2 D(P,P') \right\} \\  
&\text{s.t.} \quad \operatorname*{argmax}_i F_i(P) \neq \operatorname*{argmax}_{i'} F_{i'}(P')
\end{aligned}
\end{equation}
where $\| \mathbf{a} \|_0=\#\{i:a_i \neq 0, i=1,...,N\}$ is the $\ell_0$-norm of $\mathbf{a}$ (i.e., the number of non-zero elements in $\mathbf{a}$) and $D(P,P')$ is some distance between $P$ and $P'$.

The optimisation problem defined in Eq.~(\ref{eq_pointperturbation_1}) covers all the aforementioned conditions. In particular, the first term, $\| \mathbf{a} \|_0$ in the objective function $f(P,\mathbf{a}, E)$ imposes the quantity of selected points in the point selection process while the second term, $D(P,P')$ constrains the perceptibility of the adversarial point cloud $P'$ w.r.t. the original point cloud $P$. As will be explained later in this section, $D(P,P')$ can be defined using different distance metrics. The constraint $\operatorname*{argmax}_i F_i(P) \neq \operatorname*{argmax}_{i'} F_{i'}(P')$ ensures the generated point cloud $P'$ can fool the network $F$, i.e., $F$ would not classify $P$ and $P'$ into the same class. 

\subsubsection{Perceptibility}
There are several ways to realise the perceptibility $D(P,P')$ in Eq.~(\ref{eq_pointperturbation_1}). If we assume the correspondence between each point $\mathbf{p}_i \in P$ and its perturbed point $\mathbf{p}'_i \in P'$ defined in Eq.~(\ref{eq_adversarialpointcloud}) is maintained, then we can define $D(P,P')$ using the Euclidean distances between $\mathbf{p}_i$ and $\mathbf{p}'_i$ as,
\begin{equation} 
\label{eq_Euclideandistance}
\begin{aligned}
D_{Euclidean}(P,P') = \frac{1}{N} \sum_{i=1}^N \left(a_i \Vert \mathbf{e}_i \Vert_2 \right) \,.
\end{aligned}
\end{equation}

However, such correspondences are not always well defined, e.g., when the number of points changes in the case of point addition, making Euclidean distance not a valid choice. We further propose to use Chamfer distance and Hausdorff distance to measure perceptibility. Specifically, we can define $D(P,P')$ as
\begin{equation}
\begin{aligned}
\label{eq_Chamferdistance}
D_{Chamfer}(P, P')
=\max \bigg\{ \frac{1}{|P|} \sum_{\mathbf{p}_i \in P} \min_{\mathbf{p}'_j \in P'} \Vert \mathbf{p}_i - \mathbf{p}'_j \Vert_2,\\
\frac{1}{|P'|} \sum_{\mathbf{p}'_j \in P'} \min_{\mathbf{p}_i \in P} \Vert \mathbf{p}'_j - \mathbf{p}_i \Vert_2  \bigg\}
\end{aligned}
\end{equation}
or
\begin{equation}
\begin{aligned}
\label{eq_Hausdorffdistance}
D_{Hausdorff}(P,P')
=\max \bigg\{ \max_{\mathbf{p}_i \in P} \bigg\{ \min_{\mathbf{p}'_j \in P'} \Vert \mathbf{p}_i - \mathbf{p}'_j \Vert_2\bigg\},\\ \max_{\mathbf{p}'_j \in P'} \bigg\{ \min_{\mathbf{p}_i \in P} \Vert \mathbf{p}'_j - \mathbf{p}_i \Vert_2  \bigg\} \bigg\} \,.
\end{aligned}
\end{equation}

As shown in Eq.~(\ref{eq_Chamferdistance})-(\ref{eq_Hausdorffdistance}), both Chamfer distance and Hausdorff distance do not require the same of number of points in the point clouds $P$ and $P'$. Hence, they can be adapted easily to different point generation methods, e.g., point addition as presented in Section~\ref{sec:PointAddition}. 

\subsubsection{Relaxed Formulation}
To solve the constrained optimisation problem in Eq.~(\ref{eq_pointperturbation_1}), we convert it into an unconstrained optimisation problem using a Lagrange multiplier-like form as: 
\begin{equation} 
\label{eq_pointperturbation_3}
\begin{aligned}
&\min_{\mathbf{a}, E} f(P,\mathbf{a}, E) \\
= &\min_{\mathbf{a}, E} \left\{ \lambda_1 \| \mathbf{a} \|_0 + \lambda_2 D(P,P') + h(P') \right\}
\end{aligned}
\end{equation}
where, like~\cite{chong-first-adv-2019}, we define,
\begin{equation} 
\label{eq_classdiff}
\begin{aligned}
h(P') = \max\left\{0, F_{i^*}(P') - \max_{i^* \neq i'} F_{i'}(P')\right\}
\end{aligned}
\end{equation}
where $i^*$ is the true class label of the point set $P$.

Since the problem in Eq.~(\ref{eq_pointperturbation_3}) is NP-hard in general~\cite{Natarajan95}, we further relax it as $\ell_1$-norm optimisation~\cite{Amaldi1998,Donoho-Elad-2003} as:
\begin{equation} 
\label{eq_pointperturbation_4}
\begin{aligned}
&\min_{\mathbf{\hat{a}}, E} f(P,\mathbf{\hat{a}}, E)\\
= &\min_{\mathbf{\hat{a}}, E} \left\{ \lambda_1 \| \mathbf{\hat{a}} \|_1 + \lambda_2 D(P,P') + h(P') \right\}
\end{aligned}
\end{equation}
where $\hat{\mathbf{a}} = \left[\hat{a}_1,...,\hat{a}_N\right]^{\top} \in [0,1]^N$,  and $\| \mathbf{\hat{a}} \|_1=\sum_{i=1}^N \hat{a}_i$ is the $\ell_1$-norm of $\mathbf{\hat{a}}$.

To solve Eq.~(\ref{eq_pointperturbation_4}), we apply the iterative gradient method in~\cite{goodfellow-fgsm-iclr14,Daniel-attack-and-defense1-2019}. Since the final aim is to obtain a binary vector for $\mathbf{a}$, we randomly initialise $\hat{\mathbf{a}}$ with near-binary values, i.e., $\hat{a}_i$ is randomly set to either 0.0001 or 0.9999. Near-binary values are used to give $\hat{a}_i$ chances to turn into 1 (or 0) if $\mathbf{p}_i$ is selected (or otherwise). Finally, we achieve the final solution for $\mathbf{a}$ as,
\begin{equation} 
\label{eq_a}
\begin{aligned}
a_i = \begin{cases}
0, & \text{if } \hat{a}_i = 0 \\
1, & \text{otherwise}\,.
\end{cases}
\end{aligned}
\end{equation}

Note that we aim to find solutions for both $\mathbf{a}$ and $\mathbf{e}$. Given $\mathbf{a}$ obtained from Eq.~(\ref{eq_a}), we only consider perturbations $e_i$ if $a_i = 1$.

The optimisation problem defined in Eq.~(\ref{eq_pointperturbation_4}) formulates our proposed adversarial generation method. This formulation is general and can be adapted conveniently to other adversarial generation strategies, e.g., point addition.

\subsection{Point Addition}
\label{sec:PointAddition}

\noindent In addition to point perturbation, we can generate an adversarial example $P'$ by extending $P$ with a minimum number of additional points. We show that our proposed formulation in Eq.~(\ref{eq_pointperturbation_4}) can also be applied in this task. 
Specifically, suppose that there are no more than $K$ points added to the original point cloud $P$. We can construct a new point set $\Tilde{P}$ including all the points in $P$ and $K$ new points. These $K$ new points can be generated by randomly choosing $K$ points in $P$ and adding them to $\Tilde{P}$. We note that this way of construction of $\Tilde{P}$ does not change the perceptiblity of $P$ as $D(P, \Tilde{P})=0$ for either the Chamfer distance or Hausdorff distance used to define $D(P, \Tilde{P})$. In addition, both $P$ and $\Tilde{P}$ are treated equally by the network $F$, i.e., $\forall i, F_i(P) = F_{i}(\Tilde{P})$, as the geometric structure of the point clouds remains unchanged.

Similarly, we also construct a vector $\Tilde{\mathbf{a}}=\left[\Tilde{a}_1,...,\Tilde{a}_{N+K}\right]^{\top} \in [0,1]^{N+K}$ and a perturbation set $\Tilde{E}=\{\Tilde{\mathbf{e}}_1,...,\Tilde{\mathbf{e}}_{N+K}\}$ by extending $\mathbf{a}$ and $E$ with $K$ new elements and solve the optimisation problem in Eq.~(\ref{eq_pointperturbation_4}) with a new objective function $f(\Tilde{P}, \Tilde{\mathbf{a}}, \Tilde{E})$. The vector $\Tilde{\mathbf{a}}$ is initialised as follows, $\Tilde{a}_i$ is set to 0 for $i \in \{1,..,N\}$, and to a random value in $\{0.0001,0.9999\}$ for $i \in \{N+1,..,N+K\}$. Furthermore, during the optimisation process, we fix $\Tilde{a}_i=0$,  $\forall i=1,..,N$, i.e., original points in $P$ will not be changed. Finally, the adversarial point cloud $P'$ is obtained by including points $\Tilde{\mathbf{p}}_i \in \Tilde{P}$ such that $\Tilde{a}_i=1$.
\section{Experiments and Results}

\subsection{Experiment Setup}
\paragraph{Datasets.}

We experimented our method on  ModelNet40~\cite{Zhirong-shapenets-2014} and ScanObjectNN~\cite{Mikaela-real-world-dataset-2019} dataset. ModelNet40 is a benchmark dataset for classification of 3D CAD models. It consists of 9,843 models for training and 2,468 models for testing. We followed the experimental setup in~\cite{qi2017pointnet} to sample the surfaces of the models in ModelNet40 uniformly and normalised points into a unit cube. ScanObjectNN is an object dataset from real-world indoor scans including practical challenges such as view occlusions and object partiality. It has 15,000 objects organised in five challenging variants, e.g., objects with background, translated objects, rotated objects, and scaled objects. In our experiments, we used `OBJ\_BG', the most challenging variant including objects with background. We followed Uy et al.~\cite{Mikaela-real-world-dataset-2019} to normalise the point clouds containing background using mean and furthest point distance.

\paragraph{Implementation Details.}

We adopted PointNet~\cite{qi2017pointnet} as a test base network to conduct important experiments. To adapt with point clouds of varying sizes, we modified the max-pooling operator and batch norm accordingly. We used Adam optimiser with learning rate of 0.01. For our adversarial attack algorithm, we performed exhaustive search for the parameters in Eq.~(\ref{eq_pointperturbation_4}) and empirically set them as $\lambda_1 = 0.15$ and $\lambda_2 = 50$. Iterative gradient method in~\cite{Daniel-attack-and-defense1-2019} with 250 iterations was employed to solve Eq.~(\ref{eq_pointperturbation_4}). 

\subsection{Evaluation and Comparison}

\noindent We evaluated our method based on attack success rate, perceptibility of adversarial examples, and average number of manipulated points. The perceptibility of adversarial examples was measured using Chamfer and Hausdorff distance. We report the performance of our method on ModelNet40 and ScanObjectNN dataset in Table~\ref{table:attack_modelnet40} \&~\ref{table:attack_obj_bg} respectively. 

\begin{figure*}[t!]
\begin{subfigure}{.5\textwidth}
\includegraphics[width=\linewidth]{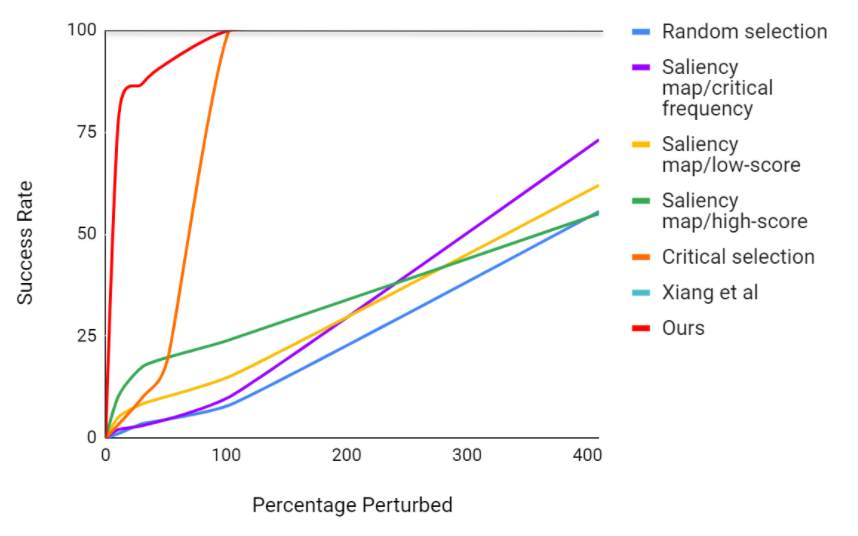}
\caption{ModelNet40}
\end{subfigure}
\begin{subfigure}{.5\textwidth}
\includegraphics[width=\linewidth]{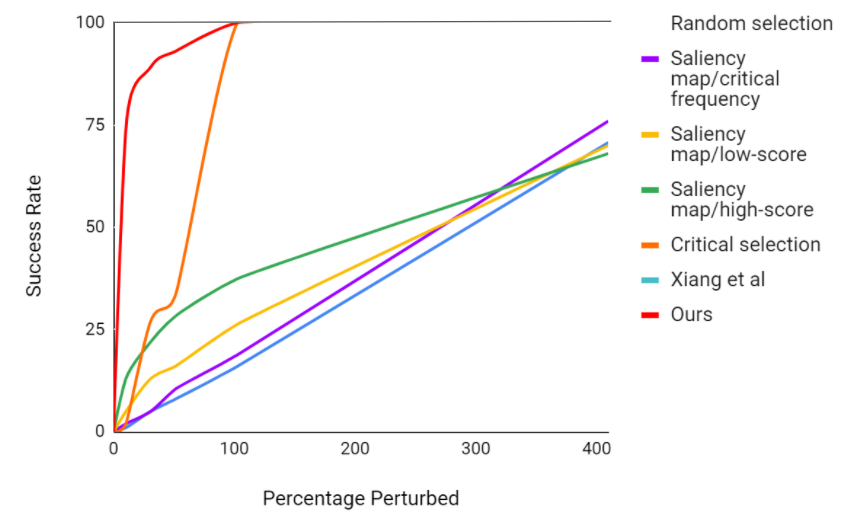}
\caption{ScanObjectNN}
\end{subfigure}
\caption{Performance trend of our adversarial attack and existing methods.}
\label{fig:graph_perturb}
\end{figure*}

\begin{figure*}[t!]
\begin{subfigure}{\textwidth}
\centering
\includegraphics[width=1.0\linewidth]{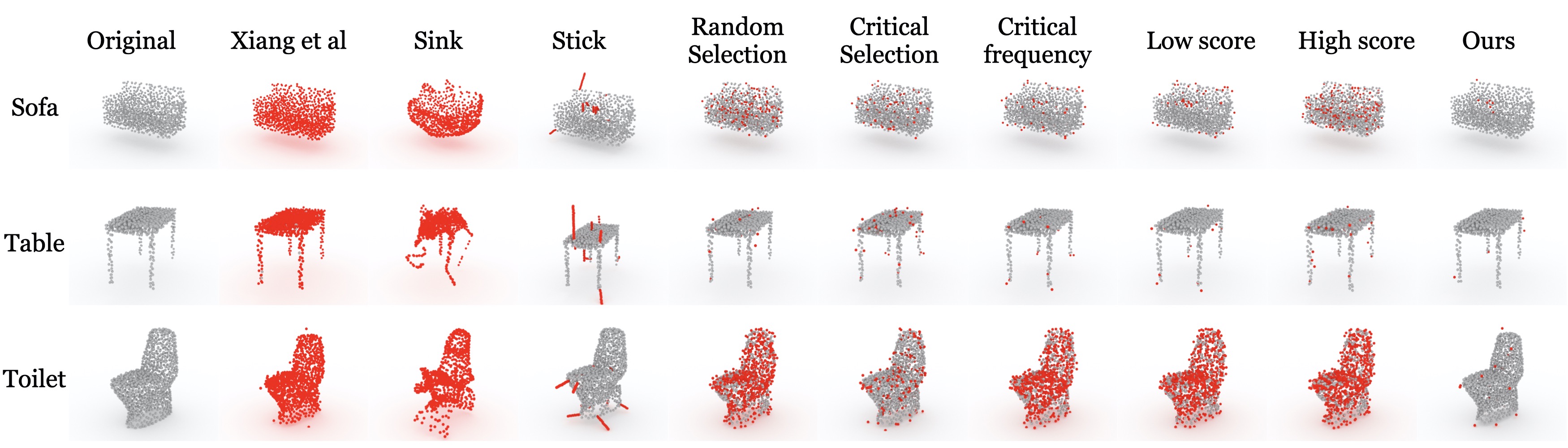}
\caption{ModelNet40}
\end{subfigure}
\begin{subfigure}{\textwidth}
\centering
\includegraphics[width=1.0\linewidth]{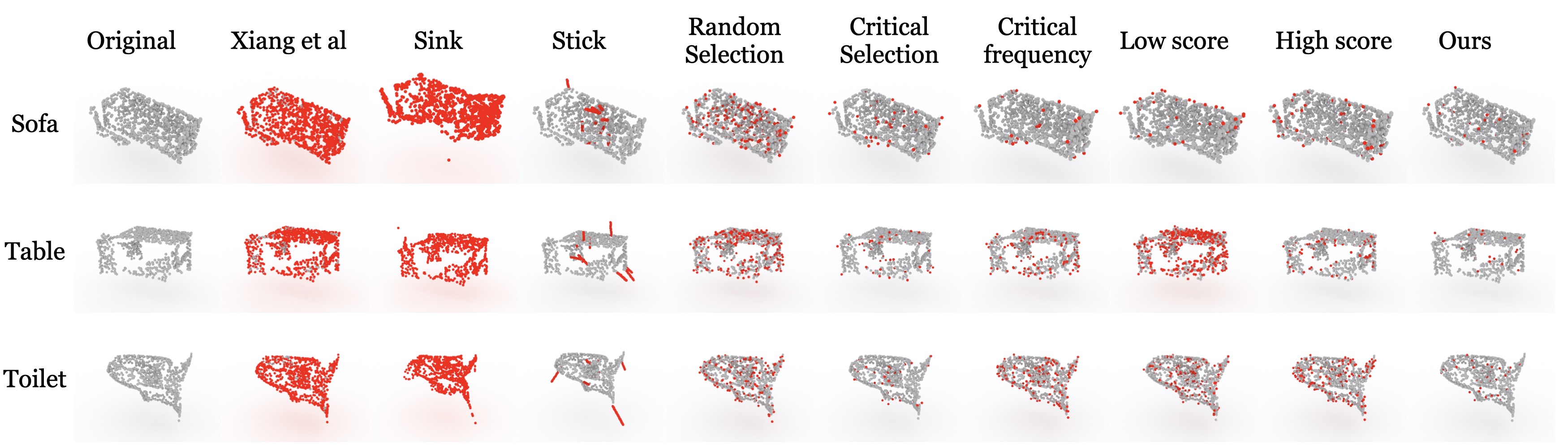}
\caption{ScanObjectNN}
\end{subfigure}
\caption{Adversarial examples generated by our method and existing methods. Red points represent perturbed points. On ScanObjectNN, our method is able to create examples that are indistinguishable from realistic noise.}
\label{fig:pert}
\end{figure*}

We also compared our method with existing methods. In particular, we tested the method by Xiang et al.~\cite{chong-first-adv-2019}, which also generated adversarial examples by point perturbation and addition. Since this method aimed for targeted attack, for fair comparison, we altered it to untargeted attack. We re-implemented the method by Liu et al.~\cite{Daniel-attack-and-defense2-2019} which made attacks in two manners: adversarial sink (i.e., pulling points towards a sink point) and adversarial stick (i.e., resampling points on added sticks). In addition, we evaluated the method by Wicker and Kwiatkowska~\cite{Matthew-Robustness-map-2019}, and by Zheng et al.~\cite{Zheng-saliency-map-2019}. Since both~\cite{Matthew-Robustness-map-2019} and~\cite{Zheng-saliency-map-2019} created adversarial examples by removing points, for fair comparison, we applied their point selection strategies but then replaced the point removal by point perturbation and point addition. In detail, the authors in~\cite{Matthew-Robustness-map-2019} selected points either randomly or from a critical point set determined by PointNet~\cite{qi2017pointnet}. In~\cite{Zheng-saliency-map-2019}, selected points were sampled from saliency maps of input point clouds in three different ways: (i) critical frequency (i.e., points frequently chosen by the max-pooling operator in PointNet~\cite{qi2017pointnet}), (ii) low-score (i.e., points having small loss gradient for a target network), and (iii) high-score (i.e., point having large loss gradient). We will further discuss the ``critical points'' in Section~\ref{sec:VulnerabilityAnalysis}.
We used 250 iterations for ~\cite{chong-first-adv-2019} and \cite{Daniel-attack-and-defense2-2019}. For other baseline methods, we perturbed the objects until a successful attack is made.

\begin{table*}
\small
\begin{center}
\begin{tabular}{l|c|cc|c}
\toprule
& Success Rate & Chamfer Distance & Hausdorff Distance & \# Points\\
\midrule
Xiang et al.~\cite{chong-first-adv-2019} & 85.9 & $1.77 \times {10^{-4}}$ & $2.38 \times {10^{-2}}$ & 967\\
Adversarial sink~\cite{Daniel-attack-and-defense2-2019} & 88.3 & $7.65 \times {10^{-3}}$ & $1.92 \times {10^{-1}}$ & 1024\\
Adversarial stick~\cite{Daniel-attack-and-defense2-2019} & 83.7 & $4.93 \times {10^{-3}}$ & $1.49 \times {10^{-1}}$ & 210\\
Random selection~\cite{Matthew-Robustness-map-2019} & 55.56 & $7.47 \times {10^{-4}}$ & $2.49 \times {10^{-3}}$ & 413\\
Critical selection~\cite{Matthew-Robustness-map-2019} & 18.99 & $1.15 \times {10^{-4}}$ & $9.39 \times {10^{-3}}$ & 50\\
Saliency map/critical frequency~\cite{Zheng-saliency-map-2019} & 63.15 & $5.72 \times {10^{-4}}$ & $2.50 \times {10^{-3}}$ & 303\\
Saliency map/low-score~\cite{Zheng-saliency-map-2019} & 55.97 & $6.47 \times {10^{-4}}$ & $2.50 \times {10^{-3}}$ & 358\\
Saliency map/high-score~\cite{Zheng-saliency-map-2019} & 58.39 & $7.52 \times {10^{-4}}$ & $2.48 \times {10^{-3}}$ & 424\\
\midrule
Ours & $\mathbf{89.38}$ & $1.55 \times {10^{-4}}$ & $1.88 \times {10^{-2}}$ & $\mathbf{36}$\\
\bottomrule 
\end{tabular}
\centerline{(a) Point Perturbation}
\newline
\newline
\begin{tabular}{l|c|cc|c}
\toprule
& Success Rate & Chamfer Distance & Hausdorff Distance & \# Points\\
\midrule 
Xiang et al.~\cite{chong-first-adv-2019} & 73.59 & $7.98 \times {10^{-3}}$ & $5.46 \times {10^{-2}}$ & 200\\
Random selection~\cite{Matthew-Robustness-map-2019} & 43.90 & $2.16 \times {10^{-4}}$ & $2.49 \times {10^{-3}}$ & 121\\
Critical selection~\cite{Matthew-Robustness-map-2019} & 47.64 & $2.05 \times {10^{-4}}$ & $2.50 \times {10^{-3}}$ & 118\\
Saliency map/critical frequency~\cite{Zheng-saliency-map-2019} & 45.13 & $2.13 \times {10^{-4}}$ & $2.49 \times {10^{-3}}$ & 118\\
Saliency map/low-score~\cite{Zheng-saliency-map-2019} & 60.96 & $1.64 \times {10^{-4}}$ & $2.50 \times {10^{-3}}$ & 89\\
Saliency map/high-score~\cite{Zheng-saliency-map-2019} & 41.06 & $2.27 \times {10^{-4}}$ & $2.49 \times {10^{-3}}$ & 128\\
\midrule 
Ours & \textbf{89.01} & $\mathbf{1.53 \times {10^{-4}}}$ & $1.98 \times {10^{-2}}$ & \textbf{38}\\
\bottomrule
\end{tabular}
\centerline{(b) Point Addition}
\caption{Attack performance to PointNet on ModelNet40.}
\label{table:attack_modelnet40}
\end{center}
\end{table*}

\begin{table*}[t]
\small
\begin{center}
\begin{tabular}{l|c|cc|c}
\toprule
& Success Rate & Chamfer Distance & Hausdorff Distance & \# Points\\
\midrule
Xiang et al.~\cite{chong-first-adv-2019} & 81.32 & $1.13 \times {10^{-4}}$ & $1.74 \times {10^{-2}}$ & 959\\
Adversarial sink~\cite{Daniel-attack-and-defense2-2019} & 78.7 & $1.37 \times {10^{-3}}$ & $9.81 \times {10^{-2}}$ & 1023\\
Adversarial stick~\cite{Daniel-attack-and-defense2-2019} & 87.5 & $5.18 \times {10^{-3}}$ & $1.67 \times {10^{-1}}$ & 210\\
Random selection~\cite{Matthew-Robustness-map-2019} & 63.72 & $6.10 \times {10^{-4}}$ & $2.50 \times {10^{-3}}$ & 340\\
Critical selection~\cite{Matthew-Robustness-map-2019} & 47.99 & $1.99 \times {10^{-4}}$ & $2.69 \times {10^{-2}}$ & 70\\
Saliency map/critical frequency~\cite{Zheng-saliency-map-2019} & 66.9 & $4.69 \times {10^{-4}}$ & $2.50 \times {10^{-3}}$ & 265\\
Saliency map/low-srop~\cite{Zheng-saliency-map-2019} & 63.81 & $5.49 \times {10^{-4}}$ & $2.50 \times {10^{-3}}$ & 306\\
Saliency map/high-srop~\cite{Zheng-saliency-map-2019} & 66.82 & $6.16 \times {10^{-4}}$ & $2.47 \times {10^{-3}}$ & 350\\
\midrule
Ours & \textbf{91.72} & $\mathbf{1.12 \times {10^{-4}}}$ & $1.15 \times {10^{-2}}$ & \textbf{34}\\
\bottomrule 
\end{tabular}
\centerline{(a) Point Perturbation}
\newline
\newline
\begin{tabular}{l|c|cc|c}
\toprule
& Success Rate & Chamfer Distance & Hausdorff Distance & \# Points\\
\midrule 
Xiang et al~\cite{chong-first-adv-2019} & 69.26 & $6.07 \times {10^{-3}}$ & $4.71 \times {10^{-2}}$ & 200\\
Random selection~\cite{Matthew-Robustness-map-2019} & 60.05 & $1.77 \times {10^{-4}}$ & $2.50 \times {10^{-3}}$ & 97\\
Critical selection~\cite{Matthew-Robustness-map-2019} & 57.44 & $1.76 \times {10^{-4}}$ & $2.50 \times {10^{-3}}$ & 98\\
Saliency map/critical frequency~\cite{Zheng-saliency-map-2019} & 59.63 & $1.79 \times {10^{-4}}$ & $2.49 \times {10^{-3}}$ & 97\\
Saliency map/low-score~\cite{Zheng-saliency-map-2019} & 60.96 & $1.64 \times {10^{-4}}$ & $2.50 \times {10^{-3}}$ & 90\\
Saliency map/high-score~\cite{Zheng-saliency-map-2019} & 57.87 & $1.87 \times {10^{-4}}$ & $2.50 \times {10^{-3}}$ & 103\\
\midrule
Ours & \textbf{90.44} & $\mathbf{1.08 \times {10^{-4}}}$ & $1.10 \times {10^{-2}}$ & \textbf{38}\\
\bottomrule
\end{tabular}
\centerline{(b) Point Addition}
\caption{Attack performance to PointNet on ScanObjectNN.}
\label{table:attack_obj_bg}
\end{center}%
\end{table*}

As shown in Table~\ref{table:attack_modelnet40} \&~\ref{table:attack_obj_bg}, compared with other methods, our method achieves the highest success rate yet lowest number of processing points for both point perturbation and point addition, and on both ModelNet40 and ScanObjectNN. Specifically, our method uses only 4\% of the total input points (1,024) to reach $>$ 89\% and $>$ 90\% of success rate on ModelNet40 and ScanObjectNN respectively. We notice that our method performs consistently (in terms of both the attack rate and the number of points) on both synthetic and real-world datasets. The adversarial sink and adversarial stick in~\cite{Daniel-attack-and-defense2-2019} respectively take the second place w.r.t. the success rate in point perturbation on ModelNet40 and ScanObjectNN. However, both of them require great numbers of points, especially the adversarial sink. The method in~\cite{chong-first-adv-2019} is ranked third for its success rate but also incurs heavy point manipulations. In addition, we observe this method often generated obvious outliers, which could be detected easily by outlier removal methods. The saliency map-based attack method~\cite{Zheng-saliency-map-2019} and the one in~\cite{Matthew-Robustness-map-2019} with random point selection require roughly $10\times$ larger point sets than our method while achieving much lower success rates. The method in~\cite{Matthew-Robustness-map-2019} with critical point selection shows relatively small number of points (though still more than our method), but the success rate is well below par. To further explore the performance trend, we plot the success rate over different numbers of points selected for point perturbation in Figure~\ref{fig:graph_perturb}. As shown in the graphs, our method significantly outperforms existing ones with a small set of points.

Experimental results also show that our method generates high-imperceptibility adversarial samples, evident by their Chamfer and Hausdorff distances to input point clouds (see Table~\ref{table:attack_modelnet40} \&~\ref{table:attack_obj_bg}). We empirically observe that Hausdorff distance results in less outlier points than Chamfer distance.
Figure~\ref{fig:pert} qualitatively compares adversarial samples generated by our method and others. As shown, our method less likely produces outliers. This is due to the use of object perceptibility and minimal point set in our formula. We observe that, adversarial examples created from real-world data (see Figure~\ref{fig:pert}(b)) are neither noticeable in perception nor distinguishable from common noise.

\begin{table*}[t]
\small
\begin{center}
\begin{tabular}{l|c|cc|c}
\toprule
& Success Rate & Chamfer Distance & Hausdorff Distance & \# Points\\
\midrule
Critical points & 88.04 & $9.53 \times {10^{-4}}$ & $8.28 \times {10^{-3}}$ & 29\\
Using all points & 91.12 & $1.15 \times {10^{-4}}$ & $8.71 \times {10^{-3}}$ & 60\\
Ours (random points) & 90.16 & $1.09 \times {10^{-4}}$ & $9.88 \times {10^{-3}}$ & 37\\
\bottomrule 
\end{tabular}
\caption{Attack performance with different initialisation strategies on ScanObjectNN.}
\label{table:various_initialization}
\end{center}
\end{table*}

\subsection{Vulnerability Analysis}
\label{sec:VulnerabilityAnalysis}

\noindent Qi et al.~\cite{qi2017pointnet} proposed a notion called ``critical points'' that characterise the shape of a point cloud. These points can be identified from the max pooling layer in PointNet~\cite{qi2017pointnet}. It is also indicated in~\cite{qi2017pointnet} that, critical points play role as an indicator for object recognition. Therefore, modification of critical points may lead to wrong classification results.

\begin{figure}[t!]
\centering
\includegraphics[width=\linewidth]{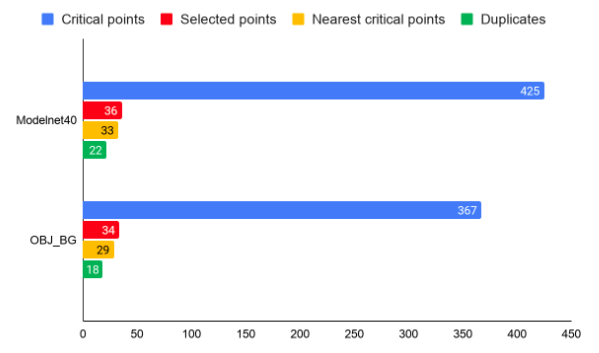}
\caption{Our selected points vs critical points.}
\label{fig:graph_Number_of_critical_and_minimum_points}
\end{figure}

\begin{figure}[t!]
\centering
\includegraphics[width=\linewidth]{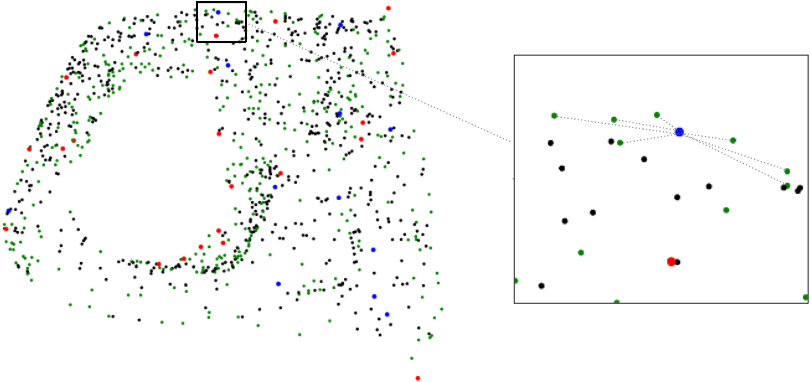}
\caption{Adversarial example of a sink. Critical points are highlighted in green. Our selected points are marked in red (if they are identical to critical points) and in blue (if they are found within the 5 nearest points of a critical point).}
\label{fig:minimum_critical_points}
\end{figure}

We found that our adversarial generation algorithm could somehow reach those critical points. To confirm this, we measured the coincidence of our selected points and PointNet's critical points. Specifically, we counted the duplicates in the two point sets, and the number of our selected points found within the 5 nearest points of a critical point. Recall that our algorithm randomly initialises the selected points (i.e., the vector $\mathbf{\hat{a}}$ in Eq.~(\ref{eq_pointperturbation_4})).  Figure~\ref{fig:graph_Number_of_critical_and_minimum_points} provides the numerical data of this experiment. It is shown that about 50\% of the selected points are identical to the critical points and 80\% of the selected points are close to the critical points. Figure~\ref{fig:minimum_critical_points} visualises our selected points and critical points. As shown, our selected points are close to critical points yet occupy a small portion, proving the capability of our method of finding compact yet vulnerable point sets.

\begin{table*}[t]
\small
\begin{center}
\begin{tabular}{l|c|cc|c}
\toprule
& Success Rate & Chamfer Distance & Hausdorff Distance & \# Points\\
\midrule
Pointnet~\cite{qi2017pointnet} & 89.38 & $1.55 \times {10^{-4}}$ & $1.88 \times {10^{-2}}$ & 36\\
Pointnet++~\cite{qi2017pointnet++} & 88.76 & $5.10 \times {10^{-4}}$ & $3.55 \times {10^{-2}}$ & 59\\
DGCNN~\cite{wang2018edgeconv} & 62.16 & $7.78 \times {10^{-4}}$ & $3.54 \times {10^{-2}}$ & 107\\
SpiderCNN~\cite{xu-spidercnn-eccv2018} & 89.92 & $4.71 \times {10^{-4}}$ & $4.16 \times {10^{-2}}$ & 57\\
PointASNL~\cite{yan2020pointasnl} & 72.16 & $2.21 \times {10^{-4}}$ & $2.03 \times {10^{-2}}$ & 45\\
\bottomrule 
\end{tabular}
\centerline{(a) Modelnet40}
\newline
\newline
\begin{tabular}{l|c|cc|c}
\toprule
& Success Rate & Chamfer Distance & Hausdorff Distance & \# Points\\
\midrule 
Pointnet~\cite{qi2017pointnet} & 91.72 & $1.12 \times {10^{-4}}$ & $1.15 \times {10^{-2}}$ & 34\\
Pointnet++~\cite{qi2017pointnet++} & 94.05 & $2.81 \times {10^{-4}}$ & $2.06 \times {10^{-2}}$ & 50\\
DGCNN~\cite{wang2018edgeconv} & 66.46 & $6.80 \times {10^{-4}}$ & $2.94 \times {10^{-2}}$ & 103\\
SpiderCNN~\cite{xu-spidercnn-eccv2018} & 86.27 & $3.23 \times {10^{-4}}$ & $2.18 \times {10^{-2}}$ & 57\\
PointASNL~\cite{yan2020pointasnl} & 55.88 & $2.07 \times {10^{-4}}$ & $1.30 \times {10^{-2}}$ & 47\\
\bottomrule
\end{tabular}
\centerline{(b) ScanObjectNN}
\caption{Attack performance to various network architectures on Modelnet40 and ScanObjectNN.}
\label{table:various_networks}
\end{center}%
\end{table*}

We also experimented our method with two additional initialisation schemes for the selected points: critical points-based initialisation and all point-based initialisation (i.e., taking all points in a point cloud to initialise the vector $\mathbf{\hat{a}}$). Table~\ref{table:various_initialization} reports the performance of various initialisation schemes. As shown in the results, the critical points-based initialisation results in the least number of points but incurs the lowest success rate. Moreover, this scheme deteriorates the perceptibility of adversarial examples (as shown in the Chamfer distances). Utilising all points for initialisation shows the opposite, i.e., more points are selected but high success rate is achieved. Our initialisation scheme compromises all the criteria, i.e., adversarial examples are created with low number of points, high attack rate, and acceptable perceptibility. In addition, random initialisation can be applied to other networks which do not support critical points.

\subsection{Attack Performance to other Architectures}

\noindent We also applied our adversarial example generation algorithm to attack existing point cloud architectures other than PointNet. Those architectures include PointNet++~\cite{qi2017pointnet++},  DGCNN~\cite{wang2018edgeconv}, SpiderCNN~\cite{xu-spidercnn-eccv2018}, and PointASNL~\cite{yan2020pointasnl}. 

We report the attack performance of our method to these network architectures in Table~\ref{table:various_networks}. Amongst all the models, DGCNN~\cite{wang2018edgeconv} is shown to be the most robust one, which requires the greatest number of points to be fooled while maintaining low attack success rate. PointNet~\cite{qi2017pointnet}, on the other hand, appears to be the most fragile model, which can be fooled easily with less points to achieve high success rate. The remaining models can be attacked by slightly different numbers of points. Table~\ref{table:various_networks} also shows that the attack performance of our method to a network architecture is consistent across both synthetic and real-world datasets.
\section{Conclusion}
\label{sec:conclusion}

\noindent In this paper, we propose a unified  formulation for minimal adversarial 3D point clouds generation that can generalise two attack strategies including point perturbation and point addition. We experimented our method on benchmark datasets, and showed that existing point cloud neural networks, e.g., PointNet, are vulnerable to attacks that perturb only 4\% of the points in a point cloud to reach more than 89\% and 90\% of success rate on synthetic and real-world data respectively. These results pose a  challenge in developing countermeasures to defend against such attacks. 

With increasingly more 3D data used in consumer devices, we envision that adversarial attack and defense for point clouds will become diverse, making this topic worthy for future research. For example, in this paper, we only investigated attacks by perturbing point cloud coordinates. For real-world point clouds, color could be another vulnerable channel for adversarial attacks. Besides, it is important to study how to create adversarial point clouds in physical world. The results of our work show that such a task could be practical as only a few percentages of an input point cloud need to be modified.

\noindent\textbf{Acknowledgment.}
Sai-Kit Yeung was partially supported by an internal grant from HKUST (R9429).

{\small
\bibliographystyle{ieee_fullname}
\bibliography{egbib}
}

\end{document}